# Fishing for Exactness *

Ted Pedersen

Department of Computer Science & Engineering
Southern Methodist University
Dallas, TX 75275
pedersen@seas.smu.edu

cmp-lg/9608010  16 Aug 1996

## Abstract

Statistical methods for automatically identifying dependent word pairs (i.e. dependent bigrams) in a corpus of natural language text have traditionally been performed using asymptotic tests of significance. This paper suggests that Fisher's exact test is a more appropriate test due to the skewed and sparse data samples typical of this problem. Both theoretical and experimental comparisons between Fisher's exact test and a variety of asymptotic tests (the t-test, Pearson's chi-square test, and Likelihood-ratio chi-square test) are presented. These comparisons show that Fisher's exact test is more reliable in identifying dependent word pairs. The usefulness of Fisher's exact test extends to other problems in statistical natural language processing as skewed and sparse data appears to be the rule in natural language. The experiment presented in this paper was performed using PROC FREQ of the SAS System.

## Introduction

Due to advances in computing power and the increasing availability of large amounts of on-line text the empirical study of human language has become an increasingly active area of research in both academic and commercial environments.

Statistical natural language processing (NLP) relies upon studying large bodies of text called corpora. These methods are useful because the unaided human mind simply cannot notice all important linguistic features, let alone rank them in order of importance, when dealing with large amounts of text.

The difficulty in studying human language based upon samples from text is that despite having billions of words on-line it is still difficult to collect large samples of certain events as many linguistic events very rarely occur. Many features of human language adhere to Zipf's Law (Zipf 1935). Informally stated, this law says that most events occur rarely and that a few very common events occur most of the time.

Statistical NLP must inevitably deal with a large number of rare events. Typical NLP data violates the large sample assumptions implicit in traditional goodness of fit tests such as Pearson's $X^2$, the Likelihood ratio $G^2$ and the t-test. When this occurs the results obtained from these tests may be in error. An alternative to these statistics is Fisher's exact test, which assigns significance by exhaustively computing all probabilities for a contingency table with fixed marginal totals.

This paper presents an experiment that compares the effectiveness of $X^2$, $G^2$, the t-test and Fisher's exact test in identifying dependent bigrams. When these results are different it is shown why Fisher's exact test gives the most reliable significance value. All of these tests can be conveniently performed using the SAS System (SAS Institute 1990).

## Lexical Relationships

A common problem in NLP is the identification of strongly associated word pairs. A bigram is any two consecutive words that occur together in a text. The frequency with which a bigram occurs throughout a text says something about the relationship between the words that make up the bigram. A *dependent bigram* is one where the two words are related in some way other than what would be expected purely by chance. Intuitively appealing examples of dependent bigrams include **major league**, **southern baptist** and **fine wine**.

The challenge in identifying dependent bigrams is that most bigrams are relatively rare regardless of the size of the text. This follows from the the distributional tendencies of individual words and bigrams as described in Zipf's Law (Zipf 1935). Zipf found that if the frequencies of the words in a large text are ordered from most to least frequent, $(f_1, f_2, \ldots, f_m)$, these frequencies roughly obey: $f_i \propto \frac{1}{i}$. The implications of Zipf's law are two-sided for statistical NLP. The good news is that a significant proportion of a corpus is made up of the most frequent words; these occur frequently enough to collect reliable statistics on them. The bad news is that there will always be a large number of words that occur just a few times.

As an example, in a 133,000 word subset of the ACL/DCI Wall Street Journal corpus (Marcus *et al.* 1993) there are 14,319 distinct words and 73,779 distinct bigrams. Of the distinct words, 48 percent of them occur only once and 80 percent of them occur five times or less. Of the distinct bigrams, 81 percent occur once and 97 percent of them occur five times or less. This data is represented graphically in Figures 1 and 2. As a result of these distributional tendencies, data samples characterizing specific bigrams are terribly skewed. This kind of data violates the *large sample assumptions* regarding the distributional characteristics of a data sample that are made by asymptotic significance tests.

---

*This research was supported by the Office of Naval Research under grant number N00014-95-1-0776.

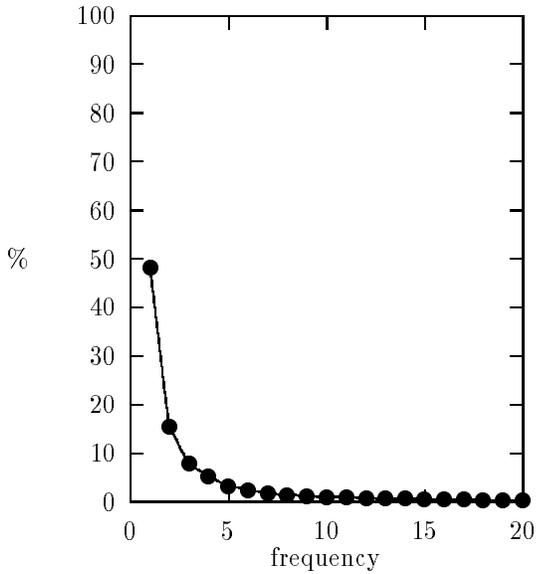

Figure 1: Distribution of single words

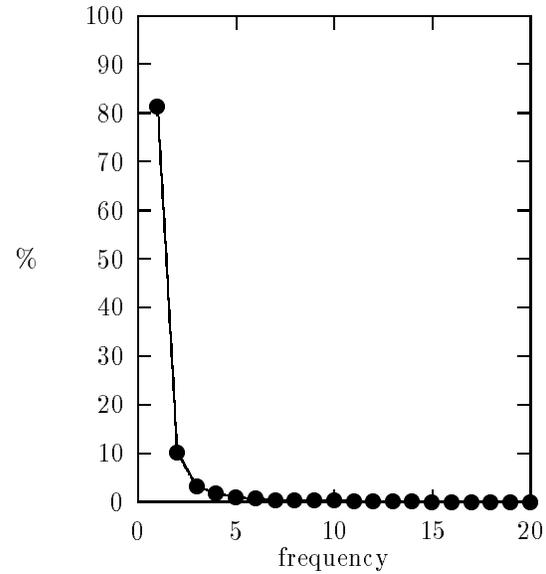

Figure 2: Distribution of bigrams

## Representation of the Data

To represent the data in terms of a statistical model, the features of each object are mapped to random variables. The relevant features of a bigram are the two words that form the bigram.

If each bigram in the data sample is characterized by two features represented by the binary variables X and Y, then each bigram will have one of four possible classifications corresponding to the possible combinations of these variable values. In this case, the data is said to be cross-classified with respect to the variables X and Y. The frequency of occurrence of these classifications can be shown in a square table having 2 rows and 2 columns. The frequency counts of each of the 4 possible data classifications in Figure 3 are denoted by $n_{11}$, $n_{12}$, $n_{21}$, and $n_{22}$.

The joint frequency distribution of X and Y is described by the counts $\{n_{ij}\}$ for the data sample represented in the contingency table. The marginal distributions of X and Y are the row and column (equation 1) totals obtained by summing the joint frequencies. The row variable is denoted $n_{i+}$ and the column variable $n_{+j}$. The subscript $+$ indicates the index over which summing has occurred.

$$n_{i+} = \sum_{j=1}^{J} n_{ij} \qquad n_{+j} = \sum_{i=1}^{I} n_{ij} \qquad (1)$$

More generally, if there are $I$ possible values for the first variable and $J$ possible values for the second variable, then the frequency of each classification can be recorded in a rectangular table having $I$ rows and $J$ columns. Each cell of this table represents one of the $I*J$ possible combinations of the variable values. Such a table is called an $I \times J$ contingency table.

|   |       | Y            |              |               |
|---|-------|--------------|--------------|---------------|
|   |       | industry     | ¬industry    | totals        |
| X | oil   | $n_{11}=$ 17 | $n_{12}=$ 229 | $n_{1+}=$ 246 |
|   | ¬oil  | $n_{21}=$ 935 | $n_{22}=$ 1381647 | $n_{2+}=$ 1382582 |
|   | totals | $n_{+1}=$952 | $n_{+2}=$1381876 | $n_{++}=$1382828 |

Figure 3: Contingency Table

As shown in Figure 3, in order to study the association (i.e., degree of dependence) between the words **oil** and **industry**, the variable X is used to denote the presence or absence of **oil** in the first position of each bigram, and Y is used to denote the presence or absence of **industry** in the second position.

## Significance Testing

In both exact and asymptotic significance testing, a probabilistic model is used to describe the distribution of the population from which the data sample was drawn. The acceptability of a potential population model is postulated as a null hypothesis and that hypothesis is tested by evaluating the *fit* of the model to the data sample. The fit is considered acceptable if the model differs from the data sample by an amount consistent with sampling variation, that is, if the value of the metric measuring the fit of the model is *statistically significant*.

The steps involved in performing a significance test are listed below and discussed in the subsections that follow. Both exact and asymptotic significance tests follow steps 1 and 2.

1. Select an appropriate sampling plan,
2. hypothesize a population model,
3. select a summary statistic to use in testing the fit of the hypothesized model to the sampled data, and
4. assess the statistical significance of the model: determine the probability that the data came from a population described by the model.

Steps 3 and 4 are more commonly associated with an asymptotic significance test. An exact test does not use a goodness of fit statistic but the notion of assessing significance still remains. The differences between the two approaches will be discussed in more detail shortly.

### Sampling Plan

In order for a significance test to yield valid results the data must be collected from the population via a random sampling plan. The sampling plan assures that each object in the data sample is selected via independent and identical trials. The sampling plan together with the population characteristics can be used to define the likelihood of selecting any particular sample.

In the experiment for this paper a multinomial sampling plan was used. In multinomial sampling the overall sample size $n_{++}$ is determined in advance and each object is randomly selected from the population to be studied. Given this, the probability of observing a particular frequency distribution $\{n_{ij}\}$ in a randomly selected data sample is shown in equation (2), where the $p_{ij}$'s are the population characteristics specifying the probability of classification $(i, j)$.

$$P(\{n_{ij}\}) = \frac{n_{++}!}{\prod_{i=1}^{I} \prod_{j=1}^{J} n_{ij}!} \prod_{i=1}^{I} \prod_{j=1}^{J} p_{ij}^{n_{ij}} \quad (2)$$

The data in Figure 3 was sampled using a multinomial sampling plan. This data is used to test the bigram **oil industry** for association. When this data was sampled, the only value that was fixed prior to the beginning of the experiment was the total sample size, $n_{++}$, which was equal to 1,382,828.

### Hypothesizing a Model

The population model used to study association between two words, where the two words are represented by the binary variables X and Y, is the model for independence between X and Y:

$$P(x, y) = P(x)P(y) \quad (3)$$

If the model for independence fits the data well as measured by its statistical significance, then one can infer from this data sample that these two words are independent in the larger population. The worse the fit, the more dependent the words are judged to be.

Using the notation introduced previously, the parameters of the model for independence between two words (i.e., the words **oil** and **industry** in Figure 3) are estimated as follows:

$$P(x) = \frac{n_{i+}}{n_{++}} \qquad P(y) = \frac{n_{+j}}{n_{++}} \quad (4)$$

In significance testing, the population model is the null hypothesis that is tested. This hypothesis can only be rejected or accepted, it can not be proven true or false with absolute certainty. The significance assigned to the hypothesis indicates how likely it is that the sample was drawn from a population specified by that model.

**Goodness of Fit Statistics** A goodness of fit statistic is used to measure how closely the counts observed in a data sample correspond to those that would be expected in a random sample drawn from a population where the null hypothesis is true.

In this section, we discuss three metrics that have been used to measure the fit of the models for association: the likelihood ratio statistic $G^2$, Pearson's $X^2$ statistic and the t-statistic. The distribution of each of these statistics can be approximated when the hypothesis is true and certain other conditions hold; they therefore can be used in asymptotic significance testing. In Fisher's exact test a goodness of fit statistic is not employed.

$G^2$ **and** $X^2$ These statistics measure the divergence of observed ($n_{ij}$) and expected ($m_{ij}$) sample counts, where the expectation is based on a hypothetical population model. These statistics can be conveniently computed using PROC FREQ of the SAS System.

The first step in calculating either $G^2$ or $X^2$ is to calculate the expected counts given that the hypothetical population model is correct. In the model for independence, maximum likelihood estimates of the expected counts are formulated as in equation (5) where $m_{ij}$ denotes the expected count in contingency table cell $(i, j)$.

$$m_{ij} = \frac{n_{i+}n_{+j}}{n_{++}} \quad (5)$$

Using this formulation, $G^2$ and $X^2$ are calculated as:

$$G^2 = 2 \sum_{i,j} n_{ij} \log \frac{n_{ij}}{m_{ij}} \qquad X^2 = \sum_{i,j} \frac{(n_{ij} - m_{ij})^2}{m_{ij}} \tag{6}$$

When the hypothetical population model is the true population model, the distribution of both $G^2$ and $X^2$ converges to $\chi^2$ as the sample size grows large (i.e., the $\chi^2$ distribution is an asymptotic approximation for the distributions of $G^2$ and $X^2$). More precisely, $X^2$ and $G^2$ are approximately $\chi^2$ distributed when the following conditions regarding the random data sample hold (Read and Cressie 1988):

1. the sample size is large,
2. the number of cells in the contingency table representation of the data is fixed and small relative to the sample size, and
3. the expected count (under the hypothetical population model) for each cell is large.

(Dunning 1993) shows that $G^2$ holds more closely to the $\chi^2$ distribution than does $X^2$ when dealing with bigram data. However, as pointed out in (Read and Cressie 1988), it is uncertain whether $G^2$ holds to the $\chi^2$ distribution when the minimum of the expected values in a table is less than 1.0. Since low expected frequencies appear to be the rule in bigram data (e.g. column $m_{11}$ in Figure 8) we suggest that the reliability of the $\chi^2$ approximation to $G^2$ could be in question.

**the t-statistic** The t-statistic (equation 7) measures the difference between the mean of a randomly drawn sample ($\overline{x}$) and the hypothesized mean for the population from which that sample was drawn ($\mu_0$). This difference is scaled by the variance of the population. When the variance of the population is unknown and the sample size is large, standard statistical techniques allow that the population variance can be estimated by the sample variance ($s^2$) which is in turn scaled by the sample size ($n$).

$$t = \frac{\overline{x} - \mu_0}{\sqrt{\frac{s^2}{n}}} \tag{7}$$

(Church et al. 1991) show how the t-statistic can be used to identify dependent bigrams. The data sample is produced through a series of Bernoulli trials that record the presence or absence of a single bigram. Given this sampling plan the sample mean is defined to be the relative frequency of the bigram ($\frac{n_{11}}{n_{++}}$) and the sample variance is roughly approximated by that same relative frequency. The t-statistic can then be rewritten as in equation 8.

$$t \approx \frac{\frac{n_{11}}{n_{++}} - \frac{n_{1+}n_{+1}}{n_{++}^2}}{\sqrt{\frac{n_{11}}{n_{++}^2}}} = \frac{n_{11} - m_{11}}{\sqrt{n_{11}}} \tag{8}$$

In the *t-test*, significance is assigned to the t-statistic using the t-distribution, which is equal to the standard normal distribution in large sample experiments. This approach to assigning significance is based on the assumption that the sample means are normally distributed. This assumption is shown to be inappropriate for bigram data in (Dunning 1993).

The formulation of (Church et al. 1991) is equivalent to a one-sample t-test. PROC TTEST of the SAS System computes a two sample t-test and was not used to compute the t-statistic values. Instead a separate data step was created to calculate the value in equation 8 and significance was assigned to that value using the PROBT function.

### Assessing Statistical Significance

If the test statistic used to evaluate a model has a known distribution when the model is correct, that distribution can be used to assign statistical significance. For $X^2$ and $G^2$ the $\chi^2$ distribution is used while the t-test uses the *t*-distribution. These serve as reliable approximations of distributions of the test statistics when certain assumptions hold. However, as has been pointed out, these assumptions are frequently violated in bigram data.

An alternative to using a significance test based on an approximate distribution is to use an *exact* significance test. In particular, for bigram data *Fisher's exact test* is recommended. This test can be performed using PROC FREQ in the SAS System.

### Fisher's Exact Test

Rather than using an asymptotic approximation of the significance of observing a particular table, Fisher's exact test (Fisher 1935) computes the significance of an observed table by exhaustively computing the probability of every table that would lead to the marginal totals that were observed in the sampled table.

The significance values obtained using Fisher's exact test are reliable regardless of the distributional characteristics of the data sample. However, when the number of comparable data samples is large, the exhaustive enumeration performed in Fisher's exact test becomes infeasible. In (Pedersen et al. 1996) an alternative test, the exact conditional test, is discussed for tables where Fisher's exact test is not a practical option.

When performing Fisher's exact test in a $2 \times 2$ contingency table the marginal totals $n_{1+}$ and $n_{+1}$ and the sample size $n_{++}$ are fixed at their observed value. Given this, the value of $n_{11}$ determines the counts in $n_{12}$, $n_{21}$ and $n_{22}$. All of the possible $2 \times 2$ tables that adhere to the fixed marginal totals are generated and the probability of each table is computed using the hypergeometric distribution.

Given that all the marginals and the sample size is fixed the hypergeometric probability of observing a particular frequency distribution $\{n_{11}, n_{12}, n_{21}, n_{22}\}$ can be computed using equation 9. Hypergeometric probabilities can be computed with the SAS System using the PROBHYPR function. PROBHYPR was used to compute the individual table probabilities the author added to the PROC FREQ output in the appendix and the data plotted in Figure 5.

$$P = \frac{1}{n_{11}!n_{12}!n_{21}!n_{22}!} * \frac{n_{1+}!n_{2+}!n_{+1}!n_{+2}!}{n_{++}!} \quad (9)$$

The original problem that Fisher used to present this test has gone down in statistical lore as the *Tea Drinker's Problem*. A woman claimed that by tasting a cup of tea with milk she could determine if the milk or tea had been added first. An experiment was designed where eight cups of tea were mixed, four cups with the milk added first and four with the tea added first. The eight cups of tea were presented to the woman in random order and she was asked to divide the 8 cups into two sets of 4, one set being those cups where milk was added first and the other those where tea was added first.

Given that the lady knew that there were 8 cups of tea and that 4 of the cups had the tea added first and 4 had the milk added first it is clear that all marginal totals should be fixed at 4 and the sample size fixed at 8. Given this there are 5 possible outcomes to the experiment ($n_{11}$ = 0, 1, 2, 3, and 4). Figure 5 shows the distributions of the hypergeometric probabilities associated with those 5 possible tables. This problem can be represented using a 2 × 2 contingency table where variable X represents the actual order of mixing and Y represents the order determined by the tea drinker. The 5 possible contingency tables and associated test values as generated by PROC FREQ are shown in the appendix.

### Interpreting the Tea Drinker's Problem

The probabilities that result from Fisher's exact test indicate how likely it is that the observed table was drawn from a population where the null hypothesis is true. In other words, the test indicates how likely it would be to randomly sample a table more supportive of the null hypothesis than the observed table. In the Tea Drinker's Problem the null hypothesis is that the tea drinker is guessing and does not really know if the milk or tea was added first. This is the hypothesis of independence and is the same hypothesis used in the test for association that identifies dependent bigrams.

Fisher's exact test can be interpreted as a one sided or two sided test. PROC FREQ shows all of the possible results: two-sided, right-sided and left-sided.

A one sided test can be either right or left sided. A right sided exact test is computed by summing the hypergeometric probabilities of all the tables with fixed marginal totals $n_{1+}$ and $n_{+1}$ whose cell count in $n_{11}$ is greater than or equal to the observed table. As an example consider the Tea Drinker's Problem where $n_{11}$ = 3. This implies that the tea drinker found 3 of the

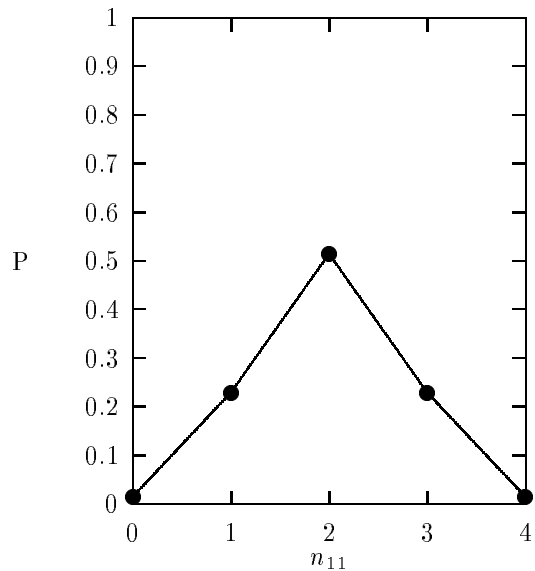

Figure 5: $n_{++} = 8, n_{1+} = 4, n_{+1} = 4$

4 cups where milk had been added first. To compute the significance of the right sided test the probabilities of the tables where $n_{11} = 3$ (.229) and $n_{11} = 4$ (.014) are summed. In this interpretation it is determined how likely it is that the tea drinker could perform more accurately in the experiment if it was repeated. A right sided test shows how likely it would be to randomly sample a table where $n_{11}$ is greater than or equal to the observed value when sampling from a population where the null hypothesis is true. The probability of being more accurate (i.e. the right sided probability) is .243 which leads to the conclusion that she is not guessing and has some idea of whether the milk or tea was added first.

The left sided test is computed in the same fashion, except that it sums the probabilities of the tables where the count in $n_{11}$ is less than or equal to the observed value. Using the same example where $n_{11} = 3$ then the probabilities of the tables where $n_{11} = 3$ (.229), $n_{11} = 2$ (.514), $n_{11} = 1$ (.229), and $n_{11} = 0$ (.014) are summed resulting in a left sided value of .986. The left sided test tells how likely it would be for the tea drinker to perform less accurately in the same experiment if it was repeated. Again, this is a fairly strong indication that the tea drinker is not simply guessing.

The two sided exact test is computed by summing the hypergeometric probabilities of all tables with the same fixed marginals but whose probabilities are less than or equal to the probability of the observed table. Consider yet again the case where $n_{11} = 3$. The probability of this table is .229. The tables that have a probability less than this are those where $n_{11} = 1$ (.229), $n_{11} = 0$ (.014) and $n_{11} = 4$ (.014). The sum of these probabilities is .486 which is the result of the two sided exact test. The question answered here is how probable would it be for the tea drinker to guess less accurately than was observed. In more general terms this is asking the question how likely would it be to randomly sample a table where the probability of observing a table where $n_{11}$ was equal to or less than the observed value when sampling from a population where the hypothesis of independence is true. On the surface this is a less convincing demonstration of the tea drinker's skill, however, it still provides reasonable evidence to support the conclusion that the tea drinker is not guessing.

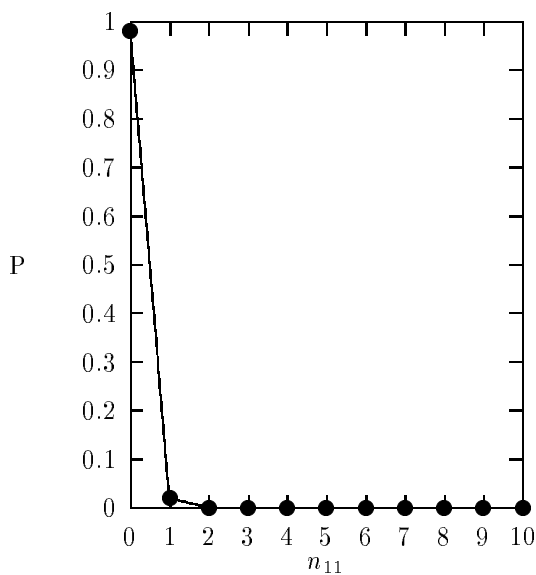

Figure 6: $n_{++} = 10000$, $n_{1+} = 20$, $n_{+1} = 10$

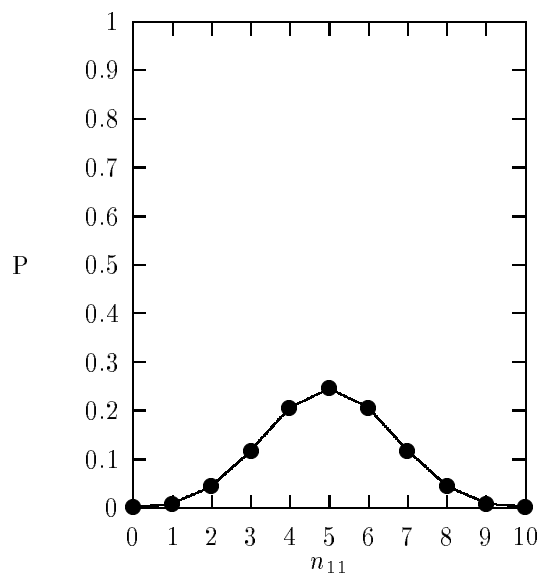

Figure 7: $n_{++} = 10000$, $n_{1+} = 5000$, $n_{+1} = 10$

## Interpreting the Bigram Experiment

The sample sizes in the bigram data are quite a bit larger than those in the Tea Drinker's Problem. In addition, the bigram data is much more skewed. However, Fisher's exact test remains a practical option since the number of possible tables is bound by the smallest marginal total (i.e, the smallest row or column total) which for bigram data is associated with the overall count of the first or second word in the bigram.

In the bigram data it would be more typical to find a sample size of 10,000 where $n_{1+} = 20$ and $n_{+1} = 10$. This implies that the row totals $n_{+1}$ and $n_{+2}$ are 10 and 9,990 respectively. In this case there are 11 possible tables where $n_{11}$ would range from 0 to 10. The distribution of hypergeometric probabilities for these 11 possible tables is shown in Figure 6.

The practical effect of the skewed distribution shown in Figure 6 is that the right sided and two sided exact test for association are equivalent. The right-sided exact test value is the probability of observing a table with $n_{11}$ greater than or equal to the observed. The two-sided exact test value is the probability of observing a table with a probability value less than or equal to the observed. These are identical since the value of P($n_{11}$) decreases as $n_{11}$ increases.

For example, if $n_{11} > 2$ the probability of observing a table with $n_{11} \geq 2$ equals .000. Thus, if $n_{11}$ is greater than 2 then there is no probability that the two words in the bigram are independent. They must be related. Notice however that if the observed count is anything but 0 a very small probability of independence is observed. Such skewed probabilities are observed for tables where $n_{++} \gg n_{1+}$ and $n_{++} \gg n_{+1}$. These sorts of tables are what was observed with the bigram data.

Notice that this was not the case in the Tea Drinker's Problem and would not be the case when the row totals are closer or equal in value. Consider the example shown in Figure 7 where $n_{++} = 10000$, $n_{1+} = 5000$ and $n_{+1} = 10$. It is easy to see that in this case $n_{1+} = n_{2+}$. In this case the distribution of the hypergeometric probabilities is symmetric and the right and two-sided tests are different.

In the test for association the marginal row totals $n_{1+}$ and $n_{2+}$ are never very close in value. $n_{1+}$ counts how many times the first word in the bigram occurs in the text overall while $n_{2+}$ is the count of all the other potential first words in the bigram. Since $n_{1+}$ will always be much less than $n_{2+}$ the distribution of the hypergeometric probabilities will always be very skewed.

In the test for association to determine bigram dependence Fisher's exact test is interpreted as a left-sided test. This shows how probable would it be to see the observed bigram a fewer number of times in another random sample from a population where the hypothesis of independence is true. If this probability is high then the words that form the bigram are dependent.

## Experiment: Test for Association[1]

There are two fundamental assumptions that underly asymptotic significance testing: (1) the data must be collected via a random sampling of the population under study, and (2) the sample must exhibit certain distributional characteristics. If either of these assumptions is not met then the inference procedure may not provide reliable results.

In this experiment, we compare the significance values computed using the t-test, the $\chi^2$ approximation to the distribution of both $G^2$ and $X^2$, and Fisher's exact test (left sided). Our data sample is a 1,382,828 word subset of the ACL/DCI Wall Street Journal corpus. We chose to characterize the associations established by the word **industry** as shown in bigrams of the form <word> **industry**. In Figure 8, we display a subset of 24 bigrams and their associated test results.

As can be seen in Figure 8, there are differences in the significance values assigned by the various tests. This indicates that the assumptions required by certain of these tests are being violated. When this occurs, the significance values assigned using Fisher's exact test should be regarded as the most reliable since there are no restrictions on the nature of the data required by this test.

Figure 8 displays the significance value assigned to the test for association between the word shown in column one and **industry**. A significance value of .0000 implies that this data shows no evidence of independence. The likelihood of having randomly selected this data sample from a population where these words were independent is zero. This is an indication of a dependent bigram. A significance value of 1.00 would indicate that there is an exact fit between the sampled data and the model for independence — there is no reason to doubt that this sample was drawn from a population in which these two words are independent. In this case the bigram is considered independent.

In this figure, we show the relative rankings of the bigrams according to their significance values. The most independent bigram is rank 1 and the most dependent bigram is rank 24. Note that the rankings defined using Fisher's exact test and the $\chi^2$ approximation to $G^2$ are identical as are the rankings as determined by the $\chi^2$ approximation to $X^2$ and the t-test. Notice further that the significance values assigned by Fisher's exact test are similar to the values as assigned by the $\chi^2$ approximation to $G^2$ for the most dependent bigrams. However, there is some variation between the significance computed for Fisher's test and $G^2$ among the more independent bigrams. This confirms the observation made by (Dunning 1993) that $G^2$ tends to overstate independence. This indicates that the asymptotic approximation of $G^2$ by the $\chi^2$ distribution is breaking down for those bigrams. In this case Fisher's test provides a more reliable significance value. The significance values assigned using the $\chi^2$ approximation to Pearson's $X^2$ and the t-test are very different from those assigned by Fisher's exact test. This indicates that neither $X^2$ nor the t-statistic is holding to its assumed asymptotic approximation.

## Conclusions

In this paper we examined recent work in identifying dependent bigrams. This work has used asymptotic significance tests when exact ones would have been more appropriate.

When asymptotic methods are used there are requirements regarding both the sampling plan and the distributional characteristics of the data that must be met. If the distributional requirements are not met, as is frequently the case in NLP, then Fisher's exact test is a viable alternative to asymptotic tests of significance. The SAS system allows for convenient computation of Fisher's exact test using PROC FREQ.

---

[1] Please contact the author at pedersen@seas.smu.edu if you would like a copy of the source code and data.

| <word> | $n_{11}$ | $m_{11}$ | exact | rank | $G^2 \sim \chi^2$ | rank | $X^2 \sim \chi^2$ | rank | T | rank |
|---|---|---|---|---|---|---|---|---|---|---|
| and | 22 | 21.14 | .8255 | 1 | .8512 | 1 | .8503 | 1 | .9958 | 1 |
| services | 1 | 0.50 | .3910 | 2 | .5293 | 2 | .4735 | 2 | .9840 | 2 |
| financial | 2 | 0.78 | .1842 | 3 | .2493 | 3 | .1673 | 3 | .9692 | 3 |
| domestic | 1 | 0.20 | .1817 | 4 | .2033 | 4 | .0740 | 4 | .9601 | 4 |
| motor | 1 | 0.15 | .1363 | 5 | .1435 | 5 | .0256 | 6 | .9502 | 6 |
| recent | 2 | 0.56 | .1073 | 6 | .1343 | 6 | .0523 | 5 | .9567 | 5 |
| instance | 1 | 0.10 | .0982 | 7 | .0971 | 7 | .0053 | 7 | .9378 | 7 |
| engineering | 1 | 0.09 | .0847 | 8 | .0815 | 8 | .0022 | 8 | .9317 | 8 |
| utility | 1 | 0.08 | .0724 | 9 | .0678 | 9 | .0007 | 10 | .9248 | 10 |
| sugar | 2 | 0.06 | .0016 | 10 | .0013 | 10 | .0000 | 16 | .8214 | 16 |
| glass | 2 | 0.04 | .0006 | 11 | .0005 | 11 | .0000 | 18 | .7746 | 18 |
| in | 7 | 21.11 | .0006 | 12 | .0003 | 12 | .0019 | 9 | .9315 | 9 |
| food | 4 | 0.29 | .0002 | 13 | .0002 | 13 | .0000 | 15 | .8475 | 15 |
| newspaper | 3 | 0.10 | .0002 | 14 | .0001 | 14 | .0000 | 17 | .8009 | 17 |
| drug | 5 | 0.47 | .0001 | 15 | .0001 | 15 | .0000 | 13 | .8531 | 13 |
| to | 9 | 27.88 | .0000 | 16 | .0000 | 16 | .0003 | 11 | .9203 | 11 |
| appliance | 2 | 0.00 | .0000 | 17 | .0000 | 17 | .0000 | 23 | .4153 | 23 |
| movie | 4 | 0.09 | .0000 | 18 | .0000 | 18 | .0000 | 19 | .7116 | 19 |
| of | 5 | 29.22 | .0000 | 19 | .0000 | 19 | .0000 | 12 | .9002 | 12 |
| futures | 8 | 0.29 | .0000 | 20 | .0000 | 21 | .0000 | 20 | .6872 | 20 |
| the | 110 | 58.95 | .0000 | 21 | .0000 | 20 | .0000 | 14 | .8524 | 14 |
| oil | 11 | 0.41 | .0000 | 22 | .0000 | 22 | .0000 | 21 | .6416 | 21 |
| airline | 17 | 0.20 | .0000 | 23 | .0000 | 24 | .0000 | 24 | .2947 | 24 |
| an | 42 | 4.03 | .0000 | 24 | .0000 | 23 | .0000 | 22 | .5965 | 22 |

Figure 8: test for association : **<word> industry**

# Appendix: PROC FREQ output for Tea Drinker's Problem

```
                          The SAS System

                        TABLE OF X BY Y

                              Y **

        Frequency|
        Expected |
        Deviation|
        Percent  |
        Row Pct  |
        Col Pct  |milk    |tea     |  Total
        ---------+--------+--------+
          milk   |    4   |    0   |    4
                 |    2   |    2   |
                 |    2   |   -2   |
                 | 50.00  |  0.00  |  50.00
                 |100.00  |  0.00  |
                 |100.00  |  0.00  |
X **    ---------+--------+--------+
          tea    |    0   |    4   |    4
                 |    2   |    2   |
                 |   -2   |    2   |
                 |  0.00  | 50.00  |  50.00
                 |  0.00  |100.00  |
                 |  0.00  |100.00  |
        ---------+--------+--------+
        Total         4        4        8
                   50.00    50.00   100.00

                 STATISTICS FOR TABLE OF X BY Y

Statistic                        DF      Value       Prob
------------------------------------------------------------
Chi-Square                        1      8.000      0.005
Likelihood Ratio Chi-Square       1     11.090      0.001
Continuity Adj. Chi-Square        1      4.500      0.034
Mantel-Haenszel Chi-Square        1      7.000      0.008
Fisher's Exact Test (Left)                          1.000
                    (Right)                         0.014
                    (2-Tail)                        0.029

P(n11 = 4)                                          0.014 **

Phi Coefficient                          1.000
Contingency Coefficient                  0.707
Cramer's V                               1.000

Sample Size = 8
WARNING: 100% of the cells have expected counts less
         than 5. Chi-Square may not be a valid test.
**: Added by the author. Not a part of PROC FREQ output
```

```
                          The SAS System

                        TABLE OF X BY Y

                              Y **

          Frequency|
          Expected |
          Deviation|
          Percent  |
          Row Pct  |
          Col Pct  |milk    |tea     |  Total
          ---------+--------+--------+
          milk     |      3 |      1 |      4
                   |      2 |      2 |
                   |      1 |     -1 |
                   |  37.50 |  12.50 |  50.00
                   |  75.00 |  25.00 |
                   |  75.00 |  25.00 |
     X**  ---------+--------+--------+
          tea      |      1 |      3 |      4
                   |      2 |      2 |
                   |     -1 |      1 |
                   |  12.50 |  37.50 |  50.00
                   |  25.00 |  75.00 |
                   |  25.00 |  75.00 |
                   ---------+--------+--------+
          Total           4        4        8
                      50.00    50.00   100.00

                STATISTICS FOR TABLE OF X BY Y

Statistic                         DF      Value       Prob
------------------------------------------------------------
Chi-Square                         1      2.000      0.157
Likelihood Ratio Chi-Square        1      2.093      0.148
Continuity Adj. Chi-Square         1      0.500      0.480
Mantel-Haenszel Chi-Square         1      1.750      0.186
Fisher's Exact Test (Left)                           0.986
                    (Right)                          0.243
                    (2-Tail)                         0.486

P(n11 = 3)                                           0.229 **

Phi Coefficient                           0.500
Contingency Coefficient                   0.447
Cramer's V                                0.500

Sample Size = 8
WARNING: 100% of the cells have expected counts less
         than 5. Chi-Square may not be a valid test.
**: Added by the author. Not a part of PROC FREQ output
```

The SAS System

TABLE OF X BY Y

```
                              Y **

         Frequency|
         Expected |
         Deviation|
         Percent  |
         Row Pct  |
         Col Pct  |milk    |tea     | Total
         ---------+--------+--------+
         milk     |    2   |    2   |    4
                  |    2   |    2   |
                  |    0   |    0   |
                  |  25.00 |  25.00 |  50.00
                  |  50.00 |  50.00 |
                  |  50.00 |  50.00 |
    X**  ---------+--------+--------+
         tea      |    2   |    2   |    4
                  |    2   |    2   |
                  |    0   |    0   |
                  |  25.00 |  25.00 |  50.00
                  |  50.00 |  50.00 |
                  |  50.00 |  50.00 |
                  ---------+--------+--------+
         Total          4        4        8
                      50.00    50.00   100.00
```

STATISTICS FOR TABLE OF X BY Y

```
Statistic                        DF      Value      Prob
------------------------------------------------------------
Chi-Square                        1      0.000      1.000
Likelihood Ratio Chi-Square       1      0.000      1.000
Continuity Adj. Chi-Square        1      0.000      1.000
Mantel-Haenszel Chi-Square        1      0.000      1.000
Fisher's Exact Test (Left)                          0.757
                    (Right)                         0.757
                    (2-Tail)                        1.000

P(n11 = 2)                                          0.514 **

Phi Coefficient                          0.000
Contingency Coefficient                  0.000
Cramer's V                               0.000

Sample Size = 8
WARNING: 100% of the cells have expected counts less
         than 5. Chi-Square may not be a valid test.
**: Added by the author. Not a part of PROC FREQ output
```

```
                            The SAS System

                          TABLE OF X BY Y

                                Y **

          Frequency|
          Expected |
          Deviation|
          Percent  |
          Row Pct  |
          Col Pct  |milk    |tea     |  Total
          ---------+--------+--------+
          milk     |    1   |    3   |    4
                   |    2   |    2   |
                   |   -1   |    1   |
                   | 12.50  | 37.50  |  50.00
                   | 25.00  | 75.00  |
                   | 25.00  | 75.00  |
    X**   ---------+--------+--------+
          tea      |    3   |    1   |    4
                   |    2   |    2   |
                   |    1   |   -1   |
                   | 37.50  | 12.50  |  50.00
                   | 75.00  | 25.00  |
                   | 75.00  | 25.00  |
                   ---------+--------+--------+
          Total          4        4        8
                       50.00    50.00   100.00

                  STATISTICS FOR TABLE OF X BY Y

Statistic                        DF      Value         Prob
------------------------------------------------------------
Chi-Square                        1      2.000        0.157
Likelihood Ratio Chi-Square       1      2.093        0.148
Continuity Adj. Chi-Square        1      0.500        0.480
Mantel-Haenszel Chi-Square        1      1.750        0.186
Fisher's Exact Test (Left)                            0.243
                    (Right)                           0.986
                    (2-Tail)                          0.486

P(n11 = 1)                                            0.229 **

Phi Coefficient                         -0.500
Contingency Coefficient                  0.447
Cramer's V                              -0.500

Sample Size = 8
WARNING: 100% of the cells have expected counts less
         than 5. Chi-Square may not be a valid test.
**: Added by the author. Not a part of PROC FREQ output
```

The SAS System

TABLE OF X BY Y

```
                         Y **

         Frequency|
         Expected |
         Deviation|
         Percent  |
         Row Pct  |
         Col Pct  |milk    |tea     |  Total
         ---------+--------+--------+
           milk   |   0    |   4    |    4
                  |   2    |   2    |
                  |  -2    |   2    |
                  |   0.00 |  50.00 |  50.00
                  |   0.00 | 100.00 |
                  |   0.00 | 100.00 |
   X**   ---------+--------+--------+
           tea    |   4    |   0    |    4
                  |   2    |   2    |
                  |   2    |  -2    |
                  |  50.00 |   0.00 |  50.00
                  | 100.00 |   0.00 |
                  | 100.00 |   0.00 |
         ---------+--------+--------+
         Total        4        4         8
                     50.00    50.00    100.00
```

STATISTICS FOR TABLE OF X BY Y

| Statistic | DF | Value | Prob |
|---|---|---|---|
| Chi-Square                   | 1 |  8.000 | 0.005 |
| Likelihood Ratio Chi-Square  | 1 | 11.090 | 0.001 |
| Continuity Adj. Chi-Square   | 1 |  4.500 | 0.034 |
| Mantel-Haenszel Chi-Square   | 1 |  7.000 | 0.008 |
| Fisher's Exact Test (Left)   |   |        | 0.014 |
|               (Right)        |   |        | 1.000 |
|               (2-Tail)       |   |        | 0.029 |
| P(n11 = 0)                   |   |        | 0.014 ** |
| Phi Coefficient              |   | -1.000 |       |
| Contingency Coefficient      |   |  0.707 |       |
| Cramer's V                   |   | -1.000 |       |

Sample Size = 8
WARNING: 100% of the cells have expected counts less
         than 5. Chi-Square may not be a valid test.
**: Added by the author. Not a part of PROC FREQ output